\documentclass[11pt,a4paper]{article}
\usepackage[hyperref]{naaclhlt2018}
\usepackage{amsmath}

\usepackage{amssymb}
\usepackage{amsfonts}
\usepackage{bbm}
\usepackage{bm}
\usepackage{booktabs}
\usepackage{color, colortbl}
\usepackage{enumitem}
\usepackage{graphicx}
\usepackage{listliketab}
\usepackage{multirow}
\usepackage{pgfplots}
\usepackage{subfig}
\usepackage{tikz}
\usepackage{times}
\usepackage{ulem}
\usepackage{url}
\usepackage{wrapfig}
\usepackage{xspace}

\usetikzlibrary{patterns}

\pgfplotsset{compat=1.11,
    /pgfplots/ybar legend/.style={
    /pgfplots/legend image code/.code={%
       \draw[##1,/tikz/.cd,yshift=-0.25em]
        (0cm,0cm) rectangle (3pt,0.8em);},
   },
}

\definecolor{forestgreen}{rgb}{0.13, 0.55, 0.13}

\aclfinalcopy 

\newcommand{\tabref}[1]{Table~\ref{tab:#1}}
\newcommand{\secref}[1]{Section~\ref{ssec:#1}}

\newcommand{\draftonly}[1]{#1}
\renewcommand{\draftonly}[1]{}
\newcommand*\samethanks[1][\value{footnote}]{\footnotemark[#1]}

\newcommand{\roy}[1]{\draftcomment{red}{roy}{#1}}

\newcommand{\omer}[1]{\draftcomment{blue}{omer}{#1}}

\newcommand{\suchin}[1]{\draftcomment{maroon}{suchin}{#1}}

\newcommand{\nascomment}[1]{\draftcomment{olive}{nas}{#1}}
\newcommand{\easy}{\textit{Easy}\xspace}
\newcommand{\hard}{\textit{Hard}\xspace}

\newcommand{\com}[1]{}
\newcommand{\resolved}[1]{}

\DeclareSymbolFont{extraup}{U}{zavm}{m}{n}
\DeclareMathSymbol{\varheart}{\mathalpha}{extraup}{86}
\makeatletter
\renewcommand*{\@fnsymbol}[1]{\bigstar}
\makeatother

\title{Annotation Artifacts in Natural Language Inference Data}

\author{ 
    Suchin Gururangan\thanks{\quad These authors contributed equally to this work.}$~~~^{\diamondsuit}$ \quad
	Swabha Swayamdipta\samethanks $~~~^\heartsuit$  \\ 
	\bf Omer Levy$^\clubsuit$ \quad
	Roy Schwartz$^{\clubsuit\spadesuit}$ \quad
	Samuel R.~Bowman $^\dagger$ \quad
	Noah A. Smith$^\clubsuit$ \\\\
	$^\diamondsuit$ Department of Linguistics, University of Washington, Seattle, WA, USA \\
	$^\heartsuit$ Language Technologies Institute, Carnegie Mellon University, Pittsburgh, PA, USA \\
	$^\clubsuit$ Paul G. Allen School of Computer Science \& Engineering, University of Washington, Seattle, WA, USA \\
	$^\spadesuit$ Allen Institute for Artificial Intelligence, Seattle, WA, USA \\
    $^\dagger$ Center for Data Science and Department of Linguistics, New York University, New York, NY, USA \\
	{\tt \{sg01,swabha,omerlevy,roysch,nasmith\}@cs.washington.edu  bowman@nyu.edu}
}

\date{}

\begin{document}
\maketitle

\resolved{\nascomment{I commented out the change to a command above, so that there would be an oxford comma. should Suchin's affiliation be UW Linguistics?  another suggested change, assuming we get 5 pages for camera ready, is to make the font in tables normal size instead of small or tiny} 
\suchin{\resolved{Yep should be linguistics! }I increased the size of the last table, but the other tables overshoot the column margins when they are normal sized. Not sure how to prevent this.}}
\resolved{\roy{any idea if we are allowed to push the text down to create a gap between affiliations and text? Alternatively, ideas on how to make the affiliations take up less space?}}
\begin{abstract}
Large-scale datasets for natural language inference are created by presenting crowd workers with a sentence (premise), and asking them to generate three new sentences (hypotheses) that it entails, contradicts, or is logically neutral with respect to. 
We show that, in a significant portion of such data, this protocol leaves clues that make it possible to identify the label by looking only at the hypothesis, without observing the premise. 
Specifically, we show that a simple text categorization model can correctly classify the hypothesis alone in about 67\% of SNLI~\cite{Bowman:2015} and 53\% of MultiNLI~\cite{Williams:2017}. 
Our analysis reveals that specific linguistic phenomena such as negation and vagueness are highly correlated with certain inference classes. 
Our findings suggest that the success of natural language inference models to date has been overestimated, and that the task remains a hard open problem. 

\end{abstract}


 \section{Introduction}

Natural language inference (NLI; also known as recognizing textual entailment, or RTE)
is a widely-studied task in natural language processing, to which many complex semantic tasks, such as question answering and text summarization, can be reduced \cite{Dagan:2006}.\resolved{\nascomment{can someone convince me this is true?  I don't see how QA or summarization reduces to classifying sentence pairs.}.}
Given a pair of sentences, a premise $p$ and a hypothesis $h$, the goal is to determine whether or not $p$ semantically entails $h$.

The problem of acquiring large amounts of labeled inference data was addressed by \citet{Bowman:2015}, who devised a method for crowdsourcing high-agreement entailment annotations en masse, creating the SNLI and later the genre-diverse MultiNLI \cite{Williams:2017} datasets.
In this process, crowd workers are presented with a premise $p$ drawn from some corpus (e.g., image captions), and are required to generate three new sentences (hypotheses) based on $p$, according to one of the following criteria:
%


%
\begin{table}[h]
\vspace{-0.5em}
\begin{tabular}{ll}
\textbf{Entailment}    & $h$ is definitely true given $p$ \\
\textbf{Neutral}       & $h$ might be true given $p$ \\
\textbf{Contradiction} & $h$ is definitely {\bf not} true given $p$ \\
\end{tabular}
\vspace{-1em}
\end{table}
\noindent
In this paper, we observe that hypotheses generated by this crowdsourcing process contain artifacts that can help a classifier detect the correct class \textit{without} ever observing the premise (\secref{artifacts}).

\begin{table*}[t]
\centering
{
\begin{tabular}{ll} \toprule
\textbf{Premise} & A woman selling bamboo sticks talking to two men on a loading dock. \\\midrule
\textbf{Entailment} & There are {\bf at least} three {\bf people} on a loading dock. \\
\textbf{Neutral} & A woman is selling bamboo sticks {\bf to help provide for her family.} \\
\textbf{Contradiction} & A woman is {\bf not} taking money for any of her sticks. \\\bottomrule
\end{tabular}
}
\caption{An instance from SNLI that illustrates the artifacts that arise from the annotation protocol. 
A common strategy for generating entailed hypotheses is to remove gender or number information. 
Neutral hypotheses are often constructed by adding a purpose clause. 
Negations are often introduced to generate contradictions.}
\label{tab:example}
\end{table*}

A closer look suggests that the observed artifacts are a product of specific annotation strategies and heuristics that crowd workers adopt. 
We find, for example, that entailed hypotheses tend to contain gender-neutral references to people, purpose clauses are a sign of neutral hypotheses, and negation is correlated with contradiction (\secref{analysis}). 
Table~\ref{tab:example} shows a single set of instances from SNLI that demonstrates all three phenomena.

We re-evaluate high-performing NLI models on the subset of examples on which our hypothesis-only classifier failed, which we consider to be ``hard'' (\secref{sota}). 
Our results show that the performance of these models on the ``hard'' subset is dramatically lower than their performance on the rest of the instances. 
This suggests that, despite recently reported progress, natural language inference remains an open problem.

\section{Annotation Artifacts are Common}
\label{ssec:artifacts}

We conjecture that the framing of the annotation task has a significant effect on the language generation choices that crowd workers make when authoring hypotheses, producing certain patterns in the data.
We call these patterns \textit{annotation artifacts}.

To determine the degree to which such artifacts exist, we train a model to predict the label of a given hypothesis \textit{without seeing the premise}.
Specifically, we use \texttt{fastText} \cite{Joulin:2017}, an off-the-shelf text classifier that models text as a bag of words and bigrams, to predict the entailment label of the hypothesis.\footnote{For MultiNLI, we additionally enabled two hyperparameters: character 4-grams, and filtering words that appeared less than 10 times in the training data.} This classifier is completely oblivious to the premise.

\tabref{results} shows that a significant portion of each test set can be correctly classified without looking at the premise, well beyond the most-frequent-class baseline.\footnote{Experiments with two other text classifiers, a logistic regression classifier with word and character $n$-gram features and a premise-oblivious version of the decomposable attention model \cite{Parikh:2016}, yielded similar results.}

Our finding demonstrates that it is possible to perform well on these datasets without modeling natural language inference.

\begin{table}[t]
\setlength{\tabcolsep}{4.5pt}

\centering
{
\begin{tabular}{lccc}
\toprule
\multirow{2}{*}{\textbf{Model}} & \multirow{2}{*}{\textbf{SNLI}} & \multicolumn{2}{c}{\textbf{MultiNLI}}  \\
 & & \textbf{Matched} & \textbf{Mismatched} \\
\midrule
majority class    & 34.3 & 35.4 & 35.2 \\
\texttt{fastText}        & \bf 67.0 & \bf 53.9 & \bf 52.3 \\
\bottomrule
\end{tabular}
}
\caption{Performance of a premise-oblivious text classifier on NLI. 
The MultiNLI benchmark contains two test sets: matched (in-domain examples) and mismatched (out-of-domain examples). 
A majority baseline is presented for reference.}
\label{tab:results}
\end{table}

\section{Characteristics of Annotation Artifacts}
\label{ssec:analysis}

In the previous section we showed that more than half (MultiNLI) or even two thirds (SNLI) of the data can be classified correctly using annotation artifacts.
A possible explanation for the formation and relative consistency of these artifacts is that crowd workers adopt  heuristics in order to generate hypotheses quickly and efficiently.
We identify some of these heuristics by conducting a shallow statistical analysis of the data, focusing on lexical choice (\secref{lexical}) and sentence length (\secref{length}).

\subsection{Lexical Choice} 
\label{ssec:lexical}

To see whether the use of certain words is indicative of the inference class, we compute the point-wise mutual information (PMI) between each word and class in the training set:
\begin{equation*}
\text{PMI} (\textit{word}, \textit{class}) = \log \frac{p(\textit{word}, \textit{class})}{p(\textit{word}, \cdot) p(\cdot, \textit{class})}
\end{equation*}
We apply add-100 smoothing to the raw statistics; 
the aggressive smoothing emphasizes word-class correlations that are highly discriminative. 
Table~\ref{tab:pmi} shows the top words affiliated with each class by PMI, along with the proportion of training sentences in each class containing each word.

Below, we elaborate on the most discriminating words for each NLI class, and suggest possible annotation heuristics that gave rise to these particular artifacts.
However, it is important to note that even the most discriminative words are not very frequent, indicating that the annotation artifacts are diverse, and that crowd workers tend to adopt multiple heuristics for generating new text. 

\begin{table*}[t]
\centering
{
\begin{tabular}{ll} \toprule
\textbf{Premise} & Two dogs are running through a field. \\\midrule
\textbf{Entailment} & There are \textbf{animals} \textbf{outdoors}. \\
\textbf{Neutral} & Some puppies are running \textbf{to catch a stick}. \\
\textbf{Contradiction} & The pets are \textbf{sitting on a couch}. \\\bottomrule
\end{tabular}
}
\caption{The example provided in the annotation guidelines for SNLI. 
Some of the observed artifacts (bold) can be potentially traced back to phenomena in this specific example.}
\label{tab:guidelines}
\end{table*}

\begin{table}
\setlength{\tabcolsep}{2pt}
\centering
\footnotesize{
\begin{tabular}{cllllll}
\toprule
 & \multicolumn{2}{c}{\textbf{Entailment}} & \multicolumn{2}{c}{\textbf{Neutral}} & \multicolumn{2}{c}{\textbf{Contradiction}} \\
\midrule
\multirow{5}{*}{\textbf{SNLI}}       & outdoors & 2.8\%     & tall & 0.7\%          & nobody & 0.1\% \\
                            & least & 0.2\%         & first & 0.6\%         & sleeping & 3.2\% \\
                            & instrument & 0.5\%   & competition & 0.7\%   & no & 1.2\% \\
                            & outside  & 8.0\%     & sad  & 0.5\%         & tv & 0.4\% \\
                            & animal & 0.7\%       & favorite & 0.4\%     & cat  & 1.3\% \\
\midrule
\multirow{5}{*}{\textbf{MNLI}}       & some & 1.6\%           & also & 1.4\%          & never & 5.0\% \\
                            & yes  & 0.1\%         & because & 4.1\%      & no & 7.6\% \\
                            & something & 0.9\%    & popular & 0.7\%      & nothing & 1.4\% \\
                            & sometimes & 0.2\%     & many & 2.2\%          & any & 4.1\% \\
                            & various & 0.1\%       & most & 1.8\%          & none & 0.1\% \\
\bottomrule
\end{tabular}
}
\caption{Top 5 words by $\text{PMI}(\text{\textit{word}}, \text{\textit{class}})$,
along with the proportion of {\it class} training samples  containing {\it word}. MultiNLI is abbreviated to MNLI.}
\label{tab:pmi}
\end{table}

\paragraph{Entailment.}
Entailed hypotheses have generic words such as \textit{animal}, \textit{instrument}, and \textit{outdoors}, which were probably chosen to generalize over more specific premise words such as \textit{dog}, \textit{guitar}, and \textit{beach}.
Other heuristics seem to replace exact numbers with approximates (\textit{some}, \textit{at least}, \textit{various}), and to remove explicit gender (\textit{human} and \textit{person} appear lower down the list).
Some artifacts are specific to the domain, such as \textit{outdoors} and \textit{outside}, which are typical of the personal photo descriptions on which SNLI was built.
Interestingly, the example from the SNLI annotation guidelines (Table~\ref{tab:guidelines}) contains both \textit{animals} and \textit{outdoors}, and also removes the number. 
This example likely primed the annotators, inducing the specific heuristics of replacing \textit{dog} with \textit{animal} and mentions of scenery with \textit{outdoors}.

\paragraph{Neutral.}
Modifiers (\textit{tall}, \textit{sad}, \textit{popular}) and superlatives (\textit{first}, \textit{favorite},\com{ \textit{many},} \textit{most}) are affiliated with the neutral class. 
These modifiers are perhaps a product of a simple strategy for introducing information that is not obviously entailed by the premise, yet plausible. 
Another formulation of neutral hypotheses seems to be through cause and purpose clauses, which increase the prevalence of discourse markers such as \textit{because}. 
Once again, we observe that the example from the SNLI annotation guidelines does just that, by adding the purpose clause \textit{to catch a stick} (Table~\ref{tab:guidelines}).

\paragraph{Contradiction.}
Negation words such as \textit{nobody}, \textit{no}, \textit{never} and \textit{nothing} 
are strong indicators of contradiction.\footnote{Similar findings were observed in the ROC story cloze annotation \cite{Schwartz:2017}.}
Other (non-negative) words appear to be part of heuristics for contradicting whatever information is displayed in the premise; \textit{sleeping} contradicts any activity, and \textit{naked} (further down the list) contradicts any description of clothing. 
The high frequency of \textit{cat} probably stems from the many dog images in the original dataset. 
\com{Another seemingly useful strategy replaces any mention of outdoor scenery with some information that implies indoors activity, such as \textit{watching TV} or \textit{sleeping}.}

\subsection{Sentence Length}
\label{ssec:length}

\begin{figure}[t]
\centering
\includegraphics[width=\columnwidth]{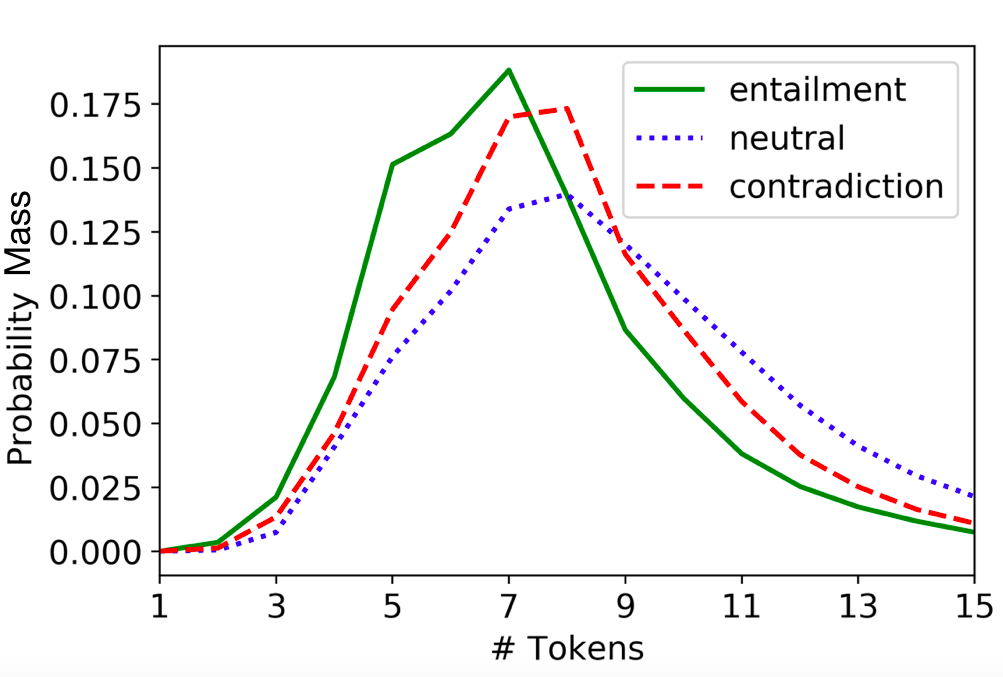}
\caption{The probability mass function of the hypothesis length in SNLI, by class.}

\label{fig:length}
\end{figure}

We observe that the number of tokens in generated hypotheses is not distributed equally among the different inference classes. 
Figure~\ref{fig:length} shows that, in SNLI, neutral hypotheses tend to be long, while entailed ones are generally shorter. 
The median length of a neutral hypothesis is 9, whereas 60\% of entailments have 7 tokens or less. 
We also observe that half of hypotheses with at least 12 tokens are neutral, while a similar portion of hypotheses of length 5 and under are entailments, making hypothesis length an effective feature. 
Length is also a discriminatory feature in MultiNLI, but is less significant, possibly due to the introduction of diverse genres.

The bias in sentence length may suggest that crowd workers created many entailed hypotheses by simply removing words from the premise. Indeed, when representing each sentence as a bag of words, 8.8\% of entailed hypotheses in SNLI are fully contained within their premise, while only 0.2\% of neutrals and contradictions exhibit the same property. MultiNLI showed similar trends.

\begin{table*}[th]
\centering
\begin{tabular}{lccccccccc}
\toprule
\multirow{2}{*}{\textbf{Model}}  & \multicolumn{3}{c}{\textbf{SNLI}} & \multicolumn{3}{c}{\textbf{MultiNLI Matched}} &  \multicolumn{3}{c}{\textbf{MultiNLI Mismatched}}   \\
 & \textit{Full} & \hard & \easy & \textit{Full} & \hard & \easy & \textit{Full} & \hard & \easy  \\
\midrule\midrule
DAM  & 84.7 & 69.4 & 92.4 & 72.0 & 55.8 & 85.3 & 72.1 & 56.2 & 85.7 \\
ESIM & 85.8 & 71.3 & 92.6 & 74.1 & 59.3 & 86.2 & 73.1 & 58.9 & 85.2 \\ 
DIIN & 86.5 & 72.7 & 93.4 & 77.0 & 64.1 & 87.6 & 76.5 & 64.4 & 86.8  \\ 
\bottomrule
\end{tabular}
\caption{Performance of high-performing NLI models on the full, \hard, and \easy NLI\com{ evaluation} test sets.
}
\label{tab:sotaresults}
\end{table*}

\section{Re-evaluating NLI Models}
\label{ssec:sota}

In Section~\ref{ssec:artifacts}, we showed that a model with no access to the premise can correctly classify many examples in both SNLI and MultiNLI, performing well above the most-frequent-class baseline.
This raises an important question about state-of-the-art NLI models: to what extent are they ``gaming'' the task by learning to detect annotation artifacts?

To answer this question, we partition each NLI test set into two subsets: examples that the premise-oblivious model classified accurately are labeled \easy, and those it could not are \hard.

We then train an NLI model on the original training sets (from either SNLI or MultiNLI),\footnote{The MultiNLI models were trained on MultiNLI data alone (as opposed to a blend of MultiNLI and SNLI data).} and evaluate on the full test set, the \hard test set, and the \easy test set.
We ran this experiment on three high-performing NLI models:
the Decomposable Attention Model (DAM; \citealp{Parikh:2016}),\footnote{
\url{github.com/allenai/allennlp}
} the Enhanced Sequential Inference Model (ESIM;  \citealp{Chen:2017}),\footnote{
\url{github.com/nyu-mll/multiNLI}
} and the Densely Interactive Inference Network (DIIN; \citealp{Gong:2017}).\footnote{
\url{goo.gl/kCeZXm}} 
All models were retrained out of the box.

Table~\ref{tab:sotaresults} shows the performance of each model on the different splits.
While the models correctly classify some \hard examples, the bulk of their success is attributed to the \easy examples. 
This result implies that the ability of NLI models to recognize textual entailment is lower than previously perceived, and that such models rely heavily on annotation artifacts in the hypothesis to make their predictions.

\resolved{\omer{Maybe add later: An additional observation from this experiment is that DIIN performs significantly better than the other two models on \hard.}\roy{True, but it also performs better than the others on the full dataset, so this is not that surprising}}

A natural question to ask is whether it is possible to select a set of NLI training and test samples which do not contain easy-to-exploit artifacts.
One  solution might be to filter  \easy examples from the training set, retaining only  \hard examples.
However, initial experiments suggest that it might not be as straightforward to eliminate annotation artifacts once the dataset has been collected.

First, after removing the \easy examples, \hard examples might not necessarily be artifact-free.
For instance, removing all contradicting samples containing the word ``no'' (a strong indicator for contradiction, see \secref{analysis}), leaves the \hard dataset with this word mostly appearing in the neutral and entailing classes,
thus creating a new artifact.
Secondly, \easy examples contain important inference phenomena (e.g. the word ``animal'' is indeed a hypernym of ``dog''), and removing these examples may hinder the model from learning such phenomena.
Importantly, artifacts do not render any particular example {\it incorrect}; they are a problem with the sample distribution, which is skewed toward certain kinds of entailment, contradiction, and neutral hypotheses.\resolved{\nascomment{added this sentence, please check}}
Therefore, a better solution might not eliminate the artifacts altogether, but rather balance them across labels.
Future strategies for reducing annotation artifacts might involve experimenting with the prompts or training given to crowd workers, e.g., to encourage a wide range of strategies, or incorporating baseline or adversarial systems that flag examples that appear to use over-represented heuristics.
We defer research on hard-to-exploit NLI datasets to future work.

\section{Discussion}

We reflect on our results and relate them to other work that also analyzes annotation artifacts in NLP datasets, drawing three main conclusions.

\paragraph{Many datasets contain annotation artifacts.}
\citet{Lai2014-cd} demonstrated that lexical features such as the presence of negation, word overlap, and hypernym relations are highly predictive of entailment classes in the SICK dataset \cite{Marelli:2014}. \citet{Chen:2016} revealed problems with the CNN/DailyMail dataset \cite{Hermann:2015} which resulted from applying automatic tools for annotation. 
\citet{Levy:2016} showed that a relation inference benchmark \cite{Zeichner:2012} is severely biased towards distributional methods, since it was created using DIRT \cite{Lin:2001}.
\citet{Schwartz:2017} and \citet{Cai:2017} showed that certain biases are prevalent in the ROC stories cloze task \cite{Mostafazadeh:2016},
which allow models trained on the endings alone, and not the story prefix, to yield state-of-the-art results. \citet{Rudinger2017SocialBI} revealed that elicited hypotheses in SNLI contain evidence of various gender, racial, religious, and aged-based stereotypes.  In parallel to this work, \citet{hypothesis-only-nli} uncovered similar annotation biases across multiple NLI datasets. Indeed, annotation artifacts are not unique to the NLI datasets, and the danger of such biases should be carefully considered when annotating new datasets.

\paragraph{Supervised models leverage annotation artifacts.}
 \citet{Levy:2015} demonstrated that supervised lexical inference models rely heavily on artifacts in the datasets, particularly the tendency of some words to serve as prototypical hypernyms. \citet{Agrawal:16,Jabri:2016,Goyal:17} all showed that state-of-the-art visual question answering~\cite{Antol:15} systems leverage annotation biases in the dataset. \citet{Cirik:18} find that complex models for referring expression recognition achieve high performance without any text input. In parallel to this work, \citet{Dasgupta2018-ry} found that the InferSent model \cite{conneau2017supervised} relies on word-level heuristics to achieve state-of-the-art performance on SNLI. These findings coincide with ours, and strongly suggest that supervised models will exploit shortcuts in the data for gaming the benchmark, if such exist.

\paragraph{Annotation artifacts inflate model performance.}
This is a corollary of the above, since large portions of the test set can be solved by relying on annotation artifacts alone. 
A similar finding by \citet{Jia:2017} showed that the performance of top question-answering models trained on SQuAD \cite{Rajpurkar:2016} drops drastically by introducing simple adversarial sentences in the evidence. 
We release the \hard SNLI and MultiNLI test sets,\footnote{SNLI: \url{goo.gl/5rQKb5}, MultiNLI matched: \url{goo.gl/abdSbi}, MultiNLI mismatched: \url{goo.gl/Cu9Gp6}\\}
and encourage the community to use them for evaluating NLI models (in addition to the original benchmarks).
We also encourage the development of additional challenging benchmarks that expose the true performance levels of state-of-the-art NLI models.

\section*{Acknowledgments}
This research was supported in part by 
the DARPA CwC program through ARO (W911NF-15-1-0543)
and a hardware gift from NVIDIA Corporation.
SB acknowledges gift support from Google and Tencent Holdings and support from Samsung Research.

\bibliography{realnli}

\begin{thebibliography}{}
\expandafter\ifx\csname natexlab\endcsname\relax\def\natexlab#1{#1}\fi

\bibitem[{Agrawal et~al.(2016)Agrawal, Batra, and Parikh}]{Agrawal:16}
Aishwarya Agrawal, Dhruv Batra, and Devi Parikh. 2016.
\newblock \href{https://aclweb.org/anthology/D16-1203}{Analyzing the behavior
  of visual question answering models}.
\newblock In {\em Proc. of EMNLP\/}.
\newblock \url{https://aclweb.org/anthology/D16-1203}.

\bibitem[{Antol et~al.(2015)Antol, Agrawal, Lu, Mitchell, Batra,
  Lawrence~Zitnick, and Parikh}]{Antol:15}
Stanislaw Antol, Aishwarya Agrawal, Jiasen Lu, Margaret Mitchell, Dhruv Batra,
  C~Lawrence~Zitnick, and Devi Parikh. 2015.
\newblock \href{https://arxiv.org/abs/1506.00278}{{VQA: Visual} question
  answering}.
\newblock In {\em Proc. of ICCV\/}.
\newblock \url{https://arxiv.org/abs/1506.00278}.

\bibitem[{Bowman et~al.(2015)Bowman, Angeli, Potts, and Manning}]{Bowman:2015}
Samuel~R. Bowman, Gabor Angeli, Christopher Potts, and Christopher~D. Manning.
  2015.
\newblock \href{https://doi.org/10.18653/v1/D15-1075}{A large annotated corpus
  for learning natural language inference}.
\newblock In {\em Proc. of EMNLP\/}.
\newblock \url{https://doi.org/10.18653/v1/D15-1075}.

\bibitem[{Cai et~al.(2017)Cai, Tu, and Gimpel}]{Cai:2017}
Zheng Cai, Lifu Tu, and Kevin Gimpel. 2017.
\newblock \href{https://doi.org/10.18653/v1/P17-2097}{Pay attention to the
  ending:strong neural baselines for the {ROC Story Cloze} task}.
\newblock In {\em Proc. of ACL\/}.
\newblock \url{https://doi.org/10.18653/v1/P17-2097}.

\bibitem[{Chen et~al.(2016)Chen, Bolton, and Manning}]{Chen:2016}
Danqi Chen, Jason Bolton, and Christopher~D. Manning. 2016.
\newblock \href{https://doi.org/10.18653/v1/P16-1223}{A thorough examination of
  the {CNN/D}aily {M}ail reading comprehension task}.
\newblock In {\em Proc. of ACL\/}.
\newblock \url{https://doi.org/10.18653/v1/P16-1223}.

\bibitem[{Chen et~al.(2017)Chen, Zhu, Ling, and Inkpen}]{Chen:2017}
Qian Chen, Xiaodan Zhu, Zhen-Hua Ling, and Diana Inkpen. 2017.
\newblock \href{https://arxiv.org/abs/1711.04289}{Natural language inference
  with external knowledge}.
\newblock {a}rXiv:1711.04289.
\newblock \url{https://arxiv.org/abs/1711.04289}.

\bibitem[{Cirik et~al.(2018)Cirik, Morency, and Berg-Kirkpatrick}]{Cirik:18}
Volkan Cirik, Louis-Philippe Morency, and Taylor Berg-Kirkpatrick. 2018.
\newblock Visual referring expression recognition: What do our systems actually
  learn?
\newblock In {\em Proc. of NAACL\/}.

\bibitem[{Conneau et~al.(2017)Conneau, Kiela, Schwenk, Barrault, and
  Bordes}]{conneau2017supervised}
Alexis Conneau, Douwe Kiela, Holger Schwenk, Loic Barrault, and Antoine Bordes.
  2017.
\newblock \href{https://doi.org/10.18653/v1/D17-1070}{Supervised learning of
  universal sentence representations from natural language inference data.}
\newblock In {\em Proc. of EMNLP\/}.
\newblock \url{https://doi.org/10.18653/v1/D17-1070}.

\bibitem[{Dagan et~al.(2006)Dagan, Glickman, and Magnini}]{Dagan:2006}
Ido Dagan, Oren Glickman, and Bernardo Magnini. 2006.
\newblock \href{https://doi.org/10.1007/11736790_9}{The {PASCAL} recognising
  textual entailment challenge}.
\newblock {\em Machine Learning Challenges\/} pages 177--190.
\newblock \url{https://doi.org/10.1007/11736790_9}.

\bibitem[{Dasgupta et~al.(2018)Dasgupta, Guo, Stuhlm{\"u}ller, Gershman, and
  Goodman}]{Dasgupta2018-ry}
Ishita Dasgupta, Demi Guo, Andreas Stuhlm{\"u}ller, Samuel~J Gershman, and
  Noah~D Goodman. 2018.
\newblock \href{https://arxiv.org/abs/1802.04302}{Evaluating compositionality
  in sentence embeddings.}
\newblock \url{https://arxiv.org/abs/1802.04302}.

\bibitem[{Gong et~al.(2018)Gong, Luo, and Zhang}]{Gong:2017}
Yichen Gong, Heng Luo, and Jian Zhang. 2018.
\newblock \href{https://arxiv.org/abs/1709.04348}{Natural language inference
  over interaction space}.
\newblock In {\em Proc. of ICLR\/}.
\newblock \url{https://arxiv.org/abs/1709.04348}.

\bibitem[{Goyal et~al.(2017)Goyal, Khot, Summers-Stay, Batra, and
  Parikh}]{Goyal:17}
Yash Goyal, Tejas Khot, Douglas Summers-Stay, Dhruv Batra, and Devi Parikh.
  2017.
\newblock \href{https://arxiv.org/abs/1612.00837}{Making the {V in VQA} matter:
  {Elevating} the role of image understanding in visual question answering}.
\newblock In {\em Proc. of CVPR\/}.
\newblock \url{https://arxiv.org/abs/1612.00837}.

\bibitem[{Hermann et~al.(2015)Hermann, Ko\v{c}isk\'{y}, Grefenstette, Espeholt,
  Kay, Suleyman, and Blunsom}]{Hermann:2015}
Karl~Moritz Hermann, Tom{\'a}\v{s} Ko\v{c}isk\'{y}, Edward Grefenstette, Lasse
  Espeholt, Will Kay, Mustafa Suleyman, and Phil Blunsom. 2015.
\newblock \href{http://dl.acm.org/citation.cfm?id=2969239.2969428}{Teaching
  machines to read and comprehend}.
\newblock In {\em Proc. of NIPS\/}.
\newblock \url{http://dl.acm.org/citation.cfm?id=2969239.2969428}.

\bibitem[{Jabri et~al.(2016)Jabri, Joulin, and van~der Maaten}]{Jabri:2016}
Allan Jabri, Armand Joulin, and Laurens van~der Maaten. 2016.
\newblock \href{https://doi.org/10.1007/978-3-319-46484-8_44}{Revisiting visual
  question answering baselines}.
\newblock In {\em Proc. of ECCV\/}.
\newblock \url{https://doi.org/10.1007/978-3-319-46484-8_44}.

\bibitem[{Jia and Liang(2017)}]{Jia:2017}
Robin Jia and Percy Liang. 2017.
\newblock \href{https://www.aclweb.org/anthology/D17-1215}{Adversarial examples
  for evaluating reading comprehension systems}.
\newblock In {\em Proc. of EMNLP\/}.
\newblock \url{https://www.aclweb.org/anthology/D17-1215}.

\bibitem[{Joulin et~al.(2017)Joulin, Grave, Bojanowski, and
  Mikolov}]{Joulin:2017}
Armand Joulin, Edouard Grave, Piotr Bojanowski, and Tomas Mikolov. 2017.
\newblock \href{https://doi.org/10.18653/v1/E17-2068}{Bag of tricks for
  efficient text classification}.
\newblock In {\em Proc. of EACL\/}.
\newblock \url{https://doi.org/10.18653/v1/E17-2068}.

\bibitem[{Lai and Hockenmaier(2014)}]{Lai2014-cd}
Alice Lai and Julia Hockenmaier. 2014.
\newblock \href{https://doi.org/10.3115/v1/S14-2055}{{Illinois-LH}: A
  denotational and distributional approach to semantics}.
\newblock In {\em Proc. of SemEval\/}.
\newblock \url{https://doi.org/10.3115/v1/S14-2055}.

\bibitem[{Levy and Dagan(2016)}]{Levy:2016}
Omer Levy and Ido Dagan. 2016.
\newblock \href{https://doi.org/10.18653/v1/P16-2041}{Annotating relation
  inference in context via question answering}.
\newblock In {\em Proc. of ACL\/}.
\newblock \url{https://doi.org/10.18653/v1/P16-2041}.

\bibitem[{Levy et~al.(2015)Levy, Remus, Biemann, and Dagan}]{Levy:2015}
Omer Levy, Steffen Remus, Chris Biemann, and Ido Dagan. 2015.
\newblock \href{https://doi.org/10.3115/v1/N15-1098}{Do supervised
  distributional methods really learn lexical inference relations?}
\newblock In {\em Proc. of NAACL\/}.
\newblock \url{https://doi.org/10.3115/v1/N15-1098}.

\bibitem[{Lin and Pantel(2001)}]{Lin:2001}
Dekang Lin and Patrick Pantel. 2001.
\newblock \href{https://doi.org/10.1017/S1351324901002765}{Discovery of
  inference rules for question-answering}.
\newblock {\em Natural Language Engineering\/} 7(4):343--360.
\newblock \url{https://doi.org/10.1017/S1351324901002765}.

\bibitem[{Marelli et~al.(2014)Marelli, Menini, Baroni, Bentivogli, Bernardi,
  and Zamparelli}]{Marelli:2014}
Marco Marelli, Stefano Menini, Marco Baroni, Luisa Bentivogli, Raffaella
  Bernardi, and Roberto Zamparelli. 2014.
\newblock \href{https://doi.org/10.3115/v1/S14-2001}{A {SICK} cure for the
  evaluation of compositional distributional semantic models.}
\newblock In {\em Proc. of LREC\/}. pages 216--223.
\newblock \url{https://doi.org/10.3115/v1/S14-2001}.

\bibitem[{Mostafazadeh et~al.(2016)Mostafazadeh, Chambers, He, Parikh, Batra,
  Vanderwende, Kohli, and Allen}]{Mostafazadeh:2016}
Nasrin Mostafazadeh, Nathanael Chambers, Xiaodong He, Devi Parikh, Dhruv Batra,
  Lucy Vanderwende, Pushmeet Kohli, and James Allen. 2016.
\newblock \href{https://doi.org/10.18653/v1/N16-1098}{A corpus and {Cloze}
  evaluation for deeper understanding of commonsense stories}.
\newblock In {\em Proc. of NAACL\/}.
\newblock \url{https://doi.org/10.18653/v1/N16-1098}.

\bibitem[{Parikh et~al.(2016)Parikh, T\"{a}ckstr\"{o}m, Das, and
  Uszkoreit}]{Parikh:2016}
Ankur Parikh, Oscar T\"{a}ckstr\"{o}m, Dipanjan Das, and Jakob Uszkoreit. 2016.
\newblock \href{https://doi.org/10.18653/v1/D16-1244}{A decomposable attention
  model for natural language inference}.
\newblock In {\em Proc. of EMNLP\/}.
\newblock \url{https://doi.org/10.18653/v1/D16-1244}.

\bibitem[{Poliak et~al.(2018)Poliak, Naradowsky, Haldar, Rudinger, and {Van
  Durme}}]{hypothesis-only-nli}
Adam Poliak, Jason Naradowsky, Aparajita Haldar, Rachel Rudinger, and Benjamin
  {Van Durme}. 2018.
\newblock Hypothesis only baselines for natural language inference.
\newblock In {\em Proc of *SEM\/}.

\bibitem[{Rajpurkar et~al.(2016)Rajpurkar, Zhang, Lopyrev, and
  Liang}]{Rajpurkar:2016}
Pranav Rajpurkar, Jian Zhang, Konstantin Lopyrev, and Percy Liang. 2016.
\newblock \href{https://doi.org/10.18653/v1/D16-1264}{{SQuAD}: 100,000+
  questions for machine comprehension of text}.
\newblock In {\em Proc. of EMNLP\/}.
\newblock \url{https://doi.org/10.18653/v1/D16-1264}.

\bibitem[{Rudinger et~al.(2017)Rudinger, May, and Durme}]{Rudinger2017SocialBI}
Rachel Rudinger, Chandler May, and Benjamin~Van Durme. 2017.
\newblock \href{http://www.aclweb.org/anthology/W17-1609}{Social bias in
  elicited natural language inferences}.
\newblock In {\em Proc. of EthNLP\/}.
\newblock \url{http://www.aclweb.org/anthology/W17-1609}.

\bibitem[{Schwartz et~al.(2017)Schwartz, Sap, Konstas, Zilles, Choi, and
  Smith}]{Schwartz:2017}
Roy Schwartz, Maarten Sap, Ioannis Konstas, Li~Zilles, Yejin Choi, and Noah~A.
  Smith. 2017.
\newblock \href{https://doi.org/10.18653/v1/K17-1004}{The effect of different
  writing tasks on linguistic style: A case study of the {ROC Story Cloze}
  task}.
\newblock In {\em Proc. of CoNLL\/}.
\newblock \url{https://doi.org/10.18653/v1/K17-1004}.

\bibitem[{Williams et~al.(2018)Williams, Nangia, and Bowman}]{Williams:2017}
Adina Williams, Nikita Nangia, and Samuel~R. Bowman. 2018.
\newblock A broad-coverage challenge corpus for sentence understanding through
  inference.
\newblock In {\em Proc. of NAACL\/}.

\bibitem[{Zeichner et~al.(2012)Zeichner, Berant, and Dagan}]{Zeichner:2012}
Naomi Zeichner, Jonathan Berant, and Ido Dagan. 2012.
\newblock \href{http://www.aclweb.org/anthology/P12-2031}{Crowdsourcing
  inference-rule evaluation}.
\newblock In {\em Proc. of ACL\/}.
\newblock \url{http://www.aclweb.org/anthology/P12-2031}.

\end{thebibliography}
\bibliographystyle{acl_natbib}

\end{document}